\documentclass[letterpaper, 10 pt, conference]{ieeeconf}  

\IEEEoverridecommandlockouts                              

\usepackage[T1]{fontenc}
\usepackage[utf8]{inputenc}
\usepackage{authblk}
\usepackage{graphicx}
\usepackage{subfigure}
\usepackage{array}
\usepackage{float}
\usepackage{float}
\usepackage{booktabs}
\usepackage{bbding}
\usepackage{etoolbox}
\usepackage{color}
\usepackage{soul}
\usepackage{algorithm}
\usepackage{algorithmic}
\usepackage{cite}
\usepackage{balance}
\usepackage{amsmath, amssymb}
\usepackage{makecell}
\usepackage{multirow}
\usepackage{hyperref}

\begin{document}

\title{\LARGE \bf
Surgical-VQLA: Transformer with Gated Vision-Language Embedding for Visual Question Localized-Answering in Robotic Surgery
}

\author{Long Bai$^{1\dagger}$, Mobarakol Islam$^{2\dagger}$, Lalithkumar Seenivasan$^3$ and Hongliang Ren$^{1,3,4*}$,\\ \textit{Senior Member, IEEE}
\thanks{$^{\dagger}$L. Bai and M. Islam are co-first authors.}%
\thanks{*The work was supported by Hong Kong Research Grants Council (RGC) Collaborative Research Fund (CRF C4026-21GF and CRF C4063-18G), and General Research Fund (GRF \#14211420 and GRF \#14216022); Shun Hing Institute of Advanced Engineering (BME-p1-21/8115064) at the CUHK; and Shenzhen-Hong Kong-Macau Technology Research Programme (Type C) Grant 202108233000303 awarded to Dr. H. Ren. M. Islam was funded by EPSRC grant [EP/W00805X/1]. We thank the CUHK Vice-Chancellor’s Ph.D. Scholarship Scheme for conference travel support. (Corresponding author: Hongliang Ren)}
\thanks{$^1$ L. Bai and H. Ren are with the Dept. of Electronic Engineering, The Chinese University of Hong Kong (CUHK), Hong Kong, China; (E-mail: b.long@ieee.org)}
\thanks{$^2$ M. Islam is with the Wellcome/EPSRC Centre for Interventional and Surgical Sciences (WEISS), University College London, UK. (E-mail: mobarakol.islam@ucl.ac.uk)}
\thanks{$^3$ L. Seenivasan and H. Ren are with Dept. of Biomedical Engineering, National University of Singapore, Singapore. (E-mail: lalithkumar\_s@u.nus.edu)}
\thanks{$^4$ H. Ren is also with Shun Hing Institute of Advanced Engineering, The Chinese University of Hong Kong (CUHK), Hong Kong 999077, China. (E-mail: hlren@ieee.org)}
}


\maketitle
\thispagestyle{empty}
\pagestyle{empty}

\begin{abstract}

Despite the availability of computer-aided simulators and recorded videos of surgical procedures, junior residents still heavily rely on experts to answer their queries. However, expert surgeons are often overloaded with clinical and academic workloads and limit their time in answering. For this purpose, we develop a surgical question-answering system to facilitate robot-assisted surgical scene and activity understanding from recorded videos. Most of the existing visual question answering (VQA) methods require an object detector and regions based feature extractor to extract visual features and fuse them with the embedded text of the question for answer generation. However, (i) surgical object detection model is scarce due to smaller datasets and lack of bounding box annotation; (ii) current fusion strategy of heterogeneous modalities like text and image is naive; (iii) the localized answering is missing, which is crucial in complex surgical scenarios. In this paper, we propose Visual Question Localized-Answering in Robotic Surgery (Surgical-VQLA) to localize the specific surgical area during the answer prediction. To deal with the fusion of the heterogeneous modalities, we design gated vision-language embedding (GVLE) to build input patches for the Language Vision Transformer (LViT) to predict the answer. To get localization, we add the detection head in parallel with the prediction head of the LViT. We also integrate generalized intersection over union (GIoU) loss to boost localization performance by preserving the accuracy of the question-answering model. We annotate two datasets of VQLA by utilizing publicly available surgical videos from EndoVis-17 and 18 of the MICCAI challenges. Our validation results suggest that Surgical-VQLA can better understand the surgical scene and localized the specific area related to the question-answering. GVLE presents an efficient language-vision embedding technique by showing superior performance over the existing benchmarks.  

\end{abstract}


\section{INTRODUCTION}
\label{sec:1}

In the absence of domain experts to answer pressing questions, the answer to "why?" could often be inferred by finding answers to "what?" and "where?". In an ideal situation, given the critical nature of the medical domain, every question on surgery and surgical procedures must be answered by expert surgeons. However, often overloaded with academic and clinical work, expert surgeons find it difficult to make time to clarify these questions~\cite{sharma2021medfusenet,seenivasan2022surgical}. To address this to an extent, recorded surgical videos are shared with the student for them to learn by observation. To improve the student’s learning experience, augmented/virtual reality-based training systems~\cite{hsieh2017vr}, automated eye tracking models~\cite{ashraf2018eye} and automated surgical skill evaluation models~\cite{lin2006towards} have also been introduced. However, these solutions still do not answer any particular questions a student might have. Their effectiveness in teaching a student relies heavily on the ability to infer from video observation and practice. Recently, MedFuseNet~\cite{sharma2021medfusenet} was proposed that performs medical visual question answering (VQA) and unfolded the possibility of developing a reliable VQA model that could supplement medical experts in answering questions from patients and students. 

\begin{figure}[h]
    \centering
    \includegraphics[width=1\linewidth, trim=0 345 480 0]{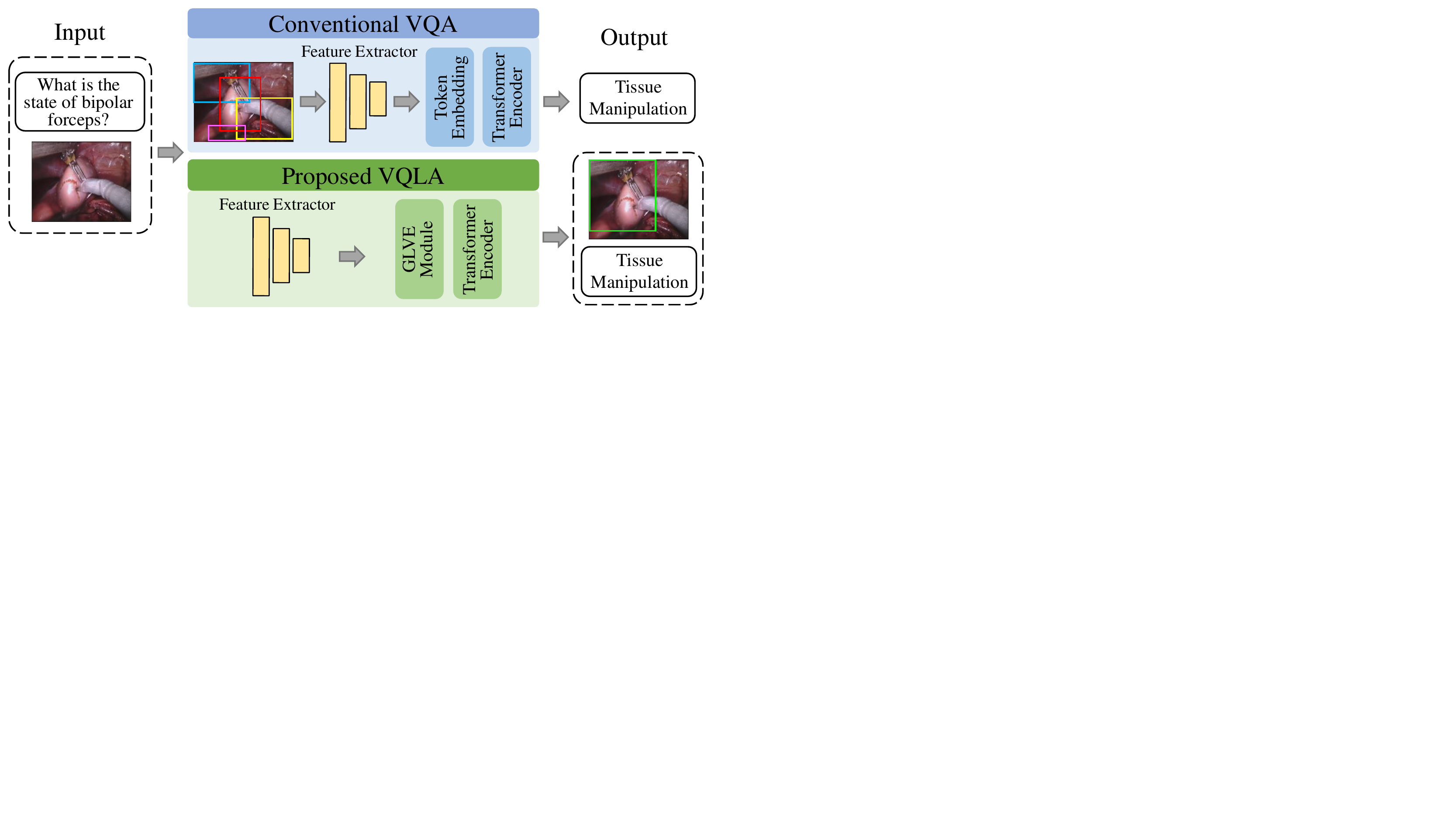}
    \caption{An overview of our proposed VQLA pipeline, against the conventional VQA tasks. Object proposals are not required in our method, and bounding box prediction can be output together with the classification results.}
    \label{fig:abstract}
\end{figure}

Very recently, Surgical-VQA~\cite{seenivasan2022surgical} has also been introduced that answers questionnaires on surgical tools, tool-tissue interactions and surgical phase based on the visual input. These two works have effectively unfolded the possibility of answering the “what?” of the questions. However, they still fail to address the “why?”. For instance, while these models could potentially answer if a patient has COVID-19 based on the X-ray scan or answer the name of tissue of interest in a surgical procedure based on the input surgical scene, it is difficult to infer the answer for “why?” from those answers. Although Surgical-VQA~\cite{seenivasan2022surgical} offers the possibility to answer the “why?” using a sentence-based open-ended VQA model, the inherent lack of annotated dataset in the medical domain still makes it difficult and time-consuming to develop a robust open-ended Surgical-VQA model. 

To outmaneuver the need for a massive annotated dataset and make it easier to infer the answer for “why?”, we propose to answer the “what?” and the “where” using a Visual Question Localized-Answering (VQLA) model in the surgical domain. In addition to answering the questions, the VQLA model also highlights the specific areas in the image related to the question and answer. This allows a better understanding of complex medical diagnoses and surgical scenes. For instance, by answering the question “what is the tissue of interest?” in a surgical procedure and indicating its location in the surgical scene, the student could then easily compare it with the surrounding tissues or counterfactual surgical procedures (surgical scenes where the tissue of interest is different) and relate the tissue (even if partially occluded) to pre-operative scan for better inference on “why?”. Localized-answer could also provide an additional advantage to students in inferring the reliability of predicted answers. For instance, if the localization is far-off from the surgical action or region of interest, it could mean that the predicted answer is less reliable. Fig.~\ref{fig:abstract} presents the overall pipeline of our proposed VQLA.

Driven by the tremendous development in deep learning and readily available enormous datasets~\cite{seenivasan2022surgical,che2022learning,bai2021influence,bai2022transformer}, significant progress has been made in developing VQA models~\cite{li2019visualbert,sheng2021human,wang2021simvlm} in the computer vision domain. Constructed using long-short term memory modules~\cite{sharma2021image,barra2021visual} or attention modules~\cite{sharma2021medfusenet,sharma2021visual}, most of these models rely heavily on object detection models, are time and resource expensive, are not end-to-end and perform a naive fusion of heterogenous features (visual and text features). Firstly, most of these models warrant employing object detection models to detect key objects in an image from which visual features are extracted. Thus, in addition to the question and answer annotations, bounding box annotations are needed to initially train the object detection model. The performance of the VQA model also relies heavily on the object detection model and a small detection error could exponentially influence the VQA model learning. Furthermore, extracting visual features only from the detected object regions and ignoring the key background features could limit the model’s global scene understanding ability \cite{seenivasan2022global} which is crucial for VQA. Secondly, as these models are trained on outputs from pre-trained object detection and feature extraction models, they are not fully end-to-end, and warrant multiple stages of training, making the overall solution sub-optimal. Thirdly, as these models are often made of multiple sub-networks (object detection, feature extraction and VQA), they are resource and time heavy, and limit usage in real-time applications. Finally, these VQA models combine the heterogenous visual and text features using naive concatenation, addition, summation, averaging or attention techniques. While these naive techniques might perform effective feature fusion for homogenous features, their performance on heterogenous features is sub-optimal as each feature hold different significance. To this extent, attentional feature fusion (AFF) and iterative attentional feature fusion (iAFF)~\cite{dai2021aff} have been recently proposed. To address the inherent limitations of using an object detection model and to perform effective heterogeneous feature fusion, we propose a detection-free Surgical VQLA model that can to trained in an end-to-end manner for localized answering based on input visual and question features. Furthermore, we propose a novel gated vision-language embedding for effective heterogeneous feature fusion and employ Generalized Intersection over Union (GIoU)~\cite{rezatofighi2019generalized} loss for robust localized-answering. 

Overall, our key contributions and findings are:
\begin{itemize}
    \item [--] We design and propose a Surgical Visual Question Localized-Answering (Surgical-VQLA\footnote{Official implementation at: \href{https://github.com/longbai1006/Surgical-VQLA}{github.com/longbai1006/Surgical-VQLA}}) model that can predict localized-answer based on a given input question and surgical scene. 
    
    \item [--] Propose a detection-free GVLE-LViT model for VQLA tasks that effectively fuse heterogeneous features (visual and text) using our novel GVLE technique. 
    \item [--] Integrate GIoU loss with cross-entropy loss and $\mathcal{L}_1$ loss to improve both the prediction and localization performance of the VQLA model. 
    \item [--] With extensive validation, we find that (i) Surgical-VQLA can localize the context even when the answer is related to surgical interaction. (ii) Our detector-free VQLA demonstrates better feature learning by avoiding computationally expensive and prone to error detection modules and facilitates the end-to-end real-time application of the surgical question localized-answering system. (iii) Proposed GVLE effectively fuses the heterogeneous modalities of visual and word embedding and outperforms existing approaches.
\end{itemize}

\section{METHODOLOGY}

\label{sec:2}

\subsection{Preliminaries}
\label{sec:2.1}

\begin{figure*}[!t]
    \centering
    \includegraphics[width=1\linewidth, trim=0 150 230 0]{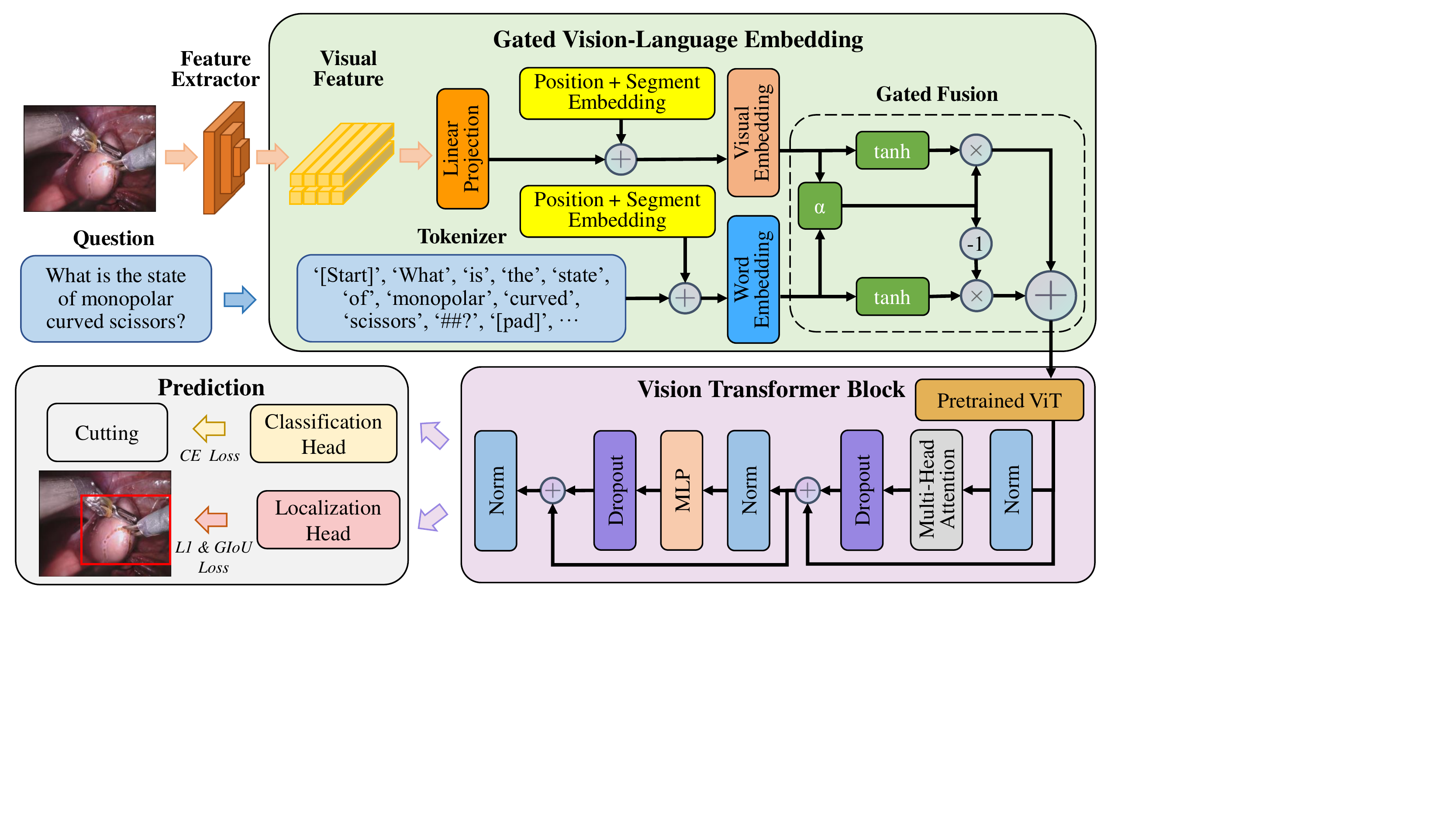}
    \caption{The proposed network architecture. The robot surgery image feeds the pre-trained feature extractor and the question feeds the customized tokenizer. The GVLE module then embeds the input features and optimizes the combination of visual and word embeddings. Fused features are propagated through the pre-trained ViT module. Finally, the answer and bounding box prediction is given by a classification head with softmax and a localization head with FFN.}
    \label{fig:overview}
\end{figure*}
\subsubsection{VisualBERT ResMLP}
\label{sec:2.1.1}
VisualBERT ResMLP~\cite{seenivasan2022surgical} is a Transformer encoder model that boosts the vision-and-language task performance of VisualBERT~\cite{li2019visualbert} with further enhancement of the input token interactions. BERT~\cite{devlin2018bert} is a Natural Language Processing model trained with subwords~\cite{wu2016google} as input. The input subwords $e$ will be mapped to a set of embeddings $e\in E$, with each embedding computed by the sum of token embedding $e_t$, segment embedding $e_s$, and position embedding $e_p$. On top of BERT~\cite{devlin2018bert}, VisualBERT~\cite{li2019visualbert} extracted visual features from object proposals to generate related visual embeddings $F$. Similarly, each embeddings $f \in F$ is the sum of visual features representation $f_v$, segment embedding $f_s$ and position embedding $f_p$. Here, position embedding is unique for each token, but segment embedding is just used to distinguish sentence and visual features. The visual and word embeddings are then combined with concatenation operation, before being sent into the multilayer Transformer, and further establish the joint inference and representation of visual and text tokens. VisualBERT ResMLP~\cite{seenivasan2022surgical} further emphasizes the token interactions based on the idea of residual MLP (ResMLP)~\cite{touvron2021resmlp} by adding cross-token and cross-channel modules in the Transformer block, which allows exchanging information between tokens.

\subsubsection{Vision Transformer}
\label{sec:2.1.2}
Vision Transformer (ViT)~\cite{dosovitskiy2020image} transfers the high performance of the Transformer~\cite{vaswani2017attention} from language tasks to vision tasks by cutting images into flattened patches. ViT~\cite{dosovitskiy2020image} is capable of capturing long-range dependency based on the self-attention mechanism, achieving notable success in vision-based tasks. After getting flattened image patches, ViT~\cite{dosovitskiy2020image} conducts patch and position embedding to preserve positional encoding information before the data go into the Transformer encoder. Finally, a multilayer perceptron (MLP) head is used for classification prediction. 

\subsection{GVLE-LViT}
\label{sec:2.2}
We develop Language-Vision Transformer (GVLE-LViT) by proposing a Gated Vision-Language Embedding (GVLE) system for efficient embedding to perform Surgical-VQLA. GVLE-LViT forms of visual feature extractor with ResNet18~\cite{he2016resnet} pre-trained on ImageNet~\cite{deng2009imagenet}, a Tokenizer, GVLE for language-vision embedding, ViT~\cite{dosovitskiy2020image} followed by a classification head and a localization head to localize spatial region while predicting the answer. Fig.~\ref{fig:overview} presents the detailed architecture of our model.

Instead of extracting visual features from object proposals like VisualBERT~\cite{li2019visualbert}, we found that pre-trained ResNet18~\cite{he2016resnet} can achieve better performance in our task. The customized tokenizer has been trained on the surgical-specific dataset for the word embeddings. The extracted visual features and word embeddings then feed the GVLE module.

\subsubsection{Gated Vision-Language Embedding (GVLE)}
\label{sec:2.2.1}
Statistical representation usually does not span modalities~\cite{srivastava2012multimodal}. Thus, the combination strategy between visual and word embeddings should be well explored.
In VisualBERT~\cite{li2019visualbert} and VisualBERT ResMLP~\cite{seenivasan2022surgical}, after conducting the sum of embeddings, respectively, visual and word embeddings are combined by the naive concatenation. At the same time, they did not consider seeking better ways of fusing representation from different sources. Inspired by Gated Multimodal Unit~\cite{arevalo2017gmu}, we borrow the idea from the flow control from recurrent neural networks. Here, the concatenation operation is replaced by a Gated Vision-Language Embedding (GVLE) module to find the best intermediate state from the visual and word embeddings. 
The right-top of Fig.~\ref{fig:overview} shows the GVLE module.
The feature embeddings of each modality are propagated through a $tanh$ activation function, which encodes the internal representation of the modality features. The gate node $\alpha$ receives the information passed from the $tanh$ activation function and decides whether the corresponding embedding information is useful. The gate is therefore used to control the weights of the obtained visual and word embeddings and constrain the model.
Therefore, the equations to combine the visual and word embeddings are as follows:
\begin{equation}
    \begin{aligned}  
        \omega &=\alpha\left(\theta_\omega \cdot\left[f\,\|\, e\,\right]\right) \\ 
        \Upsilon &=\omega * \tanh \left(\theta_f\cdot f\right) +(1-\omega) * \tanh \left(\theta_e \cdot e\right) \\ 
    \end{aligned}
    \label{equ:1}
\end{equation}

$(\theta_\omega, \theta_f, \theta_e )$ are all learnable parameters. $[\cdot\,\|\,\cdot]$ denotes concatenation operation. $f$ and $e$ represents visual and word embeddings, respectively. $\Upsilon$ is the final output of the GVLE module. The model will be able to find the best intermediate representation during training with this architecture, coupling the visual and word embeddings.
Subsequently, to fully exploit the power of pre-training, the output integrated embeddings will pass by the standard pre-trained ViT\footnote{\href{https://github.com/rwightman/pytorch-image-models}{github.com/rwightman/pytorch-image-models}}~\cite{dosovitskiy2020image} Transformer encoder and Layer-Normalization before the predication head.

\subsubsection{Prediction Head}
\label{sec:2.2.2}
The prediction head can be divided into the classification head and localization head. In the classification head, the output of the ViT~\cite{dosovitskiy2020image} block is propagated through a linear prediction layer with Softmax to achieve classification prediction.
The feed-forward network (FFN) is employed as the localization head. The FFN possesses a 3-layer perceptron with ReLU activation before a linear projection layer. 
The localization head outputs the final prediction of the normalized coordinates of the bounding box: height, width, and centre coordinates. Therefore, the system is established as an end-to-end framework.

\subsubsection{Loss Function}
\label{sec:2.2.3}
Firstly, a simple cross-entropy loss is employed as the classification loss.
Then, in the detection task, we found that the sum of $\mathcal{L}_1$ loss and GIoU loss~\cite{rezatofighi2019generalized} lead to better performance. GIoU loss~\cite{rezatofighi2019generalized} focuses on both overlapping regions and other non-overlapping regions:
\begin{equation}
    \mathcal{L}_{GIoU}=1-\left(\frac{\left|b_{g} \cap b_p\right|}{\left|b_{g} \cup b_p\right|}-\frac{\left|B\left(b_{g}, b_p\right) \backslash b_{b} \cup b_p\right|}{\left|B\left(b_{b}, b_p\right)\right|}\right)
    \label{equ:2}
\end{equation}
$b_g$ represents the ground truth bounding box, and $b_p$ denotes the predicted bounding box. $\left| \, \cdot \, \right|$ represent the area, and the operation $B$ indicate the largest box containing both $b_g$ and $b_p$.
We add the classification loss and detection loss together for the joint training. Therefore, the final loss function can be defined as:
\begin{equation}
    \begin{aligned}
        && \mathcal{L} & = \mathcal{L}_{CE} + \left(\mathcal{L}_{GIoU} +\mathcal{L}_{1}\right)
    \end{aligned}
    \label{equ:3}
\end{equation}

\section{EXPERIMENT}
\label{sec:3}

\subsection{Dataset}
\label{sec:3.1}
\subsubsection{EndoVis-18-VQLA} MICCAI Endoscopic Vision Challenge 2018~\cite{allan2020endovis18} dataset is a public dataset with 14 video sequences on robotics surgery procedures. 
We combine the bounding box on tissue-instrument interaction detection tasks~\cite{islam2020learning} and the question-answer pairs from surgical VQA classification tasks~\cite{seenivasan2022global}, generating \textbf{EndoVis-18-VQLA} with question-answer-bounding box annotations. 
Seenivasan et.al~\cite{seenivasan2022global} annotated the question-answer pairs of EndoVis-18 and make it publicly accessible\footnote{\href{https://github.com/lalithjets/surgical_vqa}{github.com/lalithjets/surgical\_vqa}}.
The answers are in single-word form with 18 distinct answer classes (1 organ, 13 tool interactions, and 4 tool locations). If the question-answer pair is only related to the organ or the tool locations, the corresponding detection bounding box will be directly given by the bounding box of the organ or the tool. 
Conversely, if the question-answer pair is related to the tissue-tool interaction, we adopt the operation $B$ in Equation~\ref{equ:2} to design a combined bounding box containing both the organ bounding box and the tool bounding box. 

We split the training and validation set by following the setup in surgical VQA classification tasks~\cite{seenivasan2022surgical}. Thus, we have 1560 frames with 9014 question-answer pairs in the training set, and 447 frames with 2769 question-answer pairs in the validation set. The EndoVis-18-VQLA dataset has been released publicly together with our official code implementation.

\subsubsection{EndoVis-17-VQLA} EndoVis-2017 Dataset~\cite{allan2019endovis17} is from the MICCAI Endoscopic Vision Challenge 2017. The original dataset contains 10 video sequences on robotics surgery scenes. To prove the generalization ability of our model, we manually select 97 frames with common tools and interactions from EndoVis-2017, and annotate the frames with question-answer-bounding box labels. Finally, we generate \textbf{EndoVis-17-VQLA} as an external validation dataset with 97 frames and 472 question-answer pairs. It is also publicly accessible with our official code implementation.

\subsection{Implementation Details}
\label{sec:3.2}
For the fair comparison, we add our prediction head in Section~\ref{sec:2.2.2} with VisualBERT\footnote{\href{https://github.com/uclanlp/visualbert}{github.com/uclanlp/visualbert}}~\cite{li2019visualbert} and VisualBERT ResMLP\footnotemark[2]~\cite{srivastava2012multimodal} and train with loss function in Equation~\ref{equ:3} to enable both classification and localization feature on these reference models.

All models are trained using the Adam optimizer~\cite{kingma2014adam} for $80$ epochs with a batch size of $64$ and a learning rate of $1 \times 10^{-5}$.
The models are trained on EndoVis-18-VQLA training set, validated on both EndoVis-18-VQLA validation set and EndoVis-17-VQLA, an external validation dataset. All experiments are implemented by Python PyTorch framework, and conducted on a server with NVIDIA RTX 3090 GPU and Intel Core i9-10980XE CPU.

\subsection{Results}
\label{sec:3.3}

\begin{table*}[!h]
    \renewcommand\arraystretch{0.4}
	\caption{
	    Comparison experiments of our GVLE-LViT model, against VisualBERT~\cite{li2019visualbert} and VisualBERT ResMLP~\cite{seenivasan2022surgical} based model. RN denotes ResNet.
	}
	\centering
	\label{tab:1}  
	\setlength{\tabcolsep}{1.5mm}{
	\begin{tabular}{c|p{1.2cm}<{\centering} p{1.2cm}<{\centering} p{1.2cm}<{\centering}|p{1.0cm}<{\centering} p{1.0cm}<{\centering} p{1.0cm}<{\centering} | p{1.0cm}<{\centering} p{1.0cm}<{\centering} p{1.0cm}<{\centering}}
		\noalign{\smallskip}\hline\noalign{\smallskip}\noalign{\smallskip}	
        & \multicolumn{3}{c|}{Visual Feature} & \multicolumn{3}{c}{\textbf{EndoVis-18-VQLA}} & \multicolumn{3}{|c}{\textbf{EndoVis-17-VQLA}  } \\\noalign{\smallskip}\cline{2-10} \noalign{\smallskip}	
        \multicolumn{1}{c|}{\makecell[c]{Model \\ \textcolor{white}{1}}} & Detection & \multicolumn{1}{c}{\makecell[c]{Feature \\ Extraction}} & FPS & Acc & F-Score & mIoU & Acc & F-Score & mIoU \\\noalign{\smallskip}\hline\noalign{\smallskip}
		VisualBERT~\cite{li2019visualbert} & & & & 0.5883 & 0.3012 & 0.7383 & 0.4428 & \textbf{0.3844} & 0.7094\\
		\noalign{\smallskip}
		VisualBERT ResMLP~\cite{seenivasan2022surgical} & FRCNN~\cite{ren2015faster} & RN~\cite{he2016resnet} & 18.09 & 0.6049 & 0.3045 & 0.7287 & 0.4258 & 0.3702 & 0.6803\\
        \noalign{\smallskip}
        GVLE-LViT (Ours) & & & & 0.6079 & \textbf{0.3677} & 0.7122 & 0.4407 & 0.3273 & 0.6852\\\noalign{\smallskip}\hline\noalign{\smallskip}
		VisualBERT~\cite{li2019visualbert} & & & & 0.6215 & 0.3320 & 0.7356 & 0.3898 & 0.3169 & 0.7105 \\
		\noalign{\smallskip}
		VisualBERT ResMLP~\cite{seenivasan2022surgical} & $\times$ & RN~\cite{he2016resnet} & 150.60 & 0.6320 & 0.3311 & 0.7501 & 0.4195 & 0.3316 & 0.7035 \\
        \noalign{\smallskip}
        GVLE-LViT (Ours) & & & & \textbf{0.6659} & 0.3614 & \textbf{0.7625} & \textbf{0.4576} & 0.2489 & \textbf{0.7275} \\
		\noalign{\smallskip}\hline
	\end{tabular}}
\end{table*}

\begin{figure*}[!t]
    \centering
    \includegraphics[width=1\linewidth, trim=0 95 0 -20]{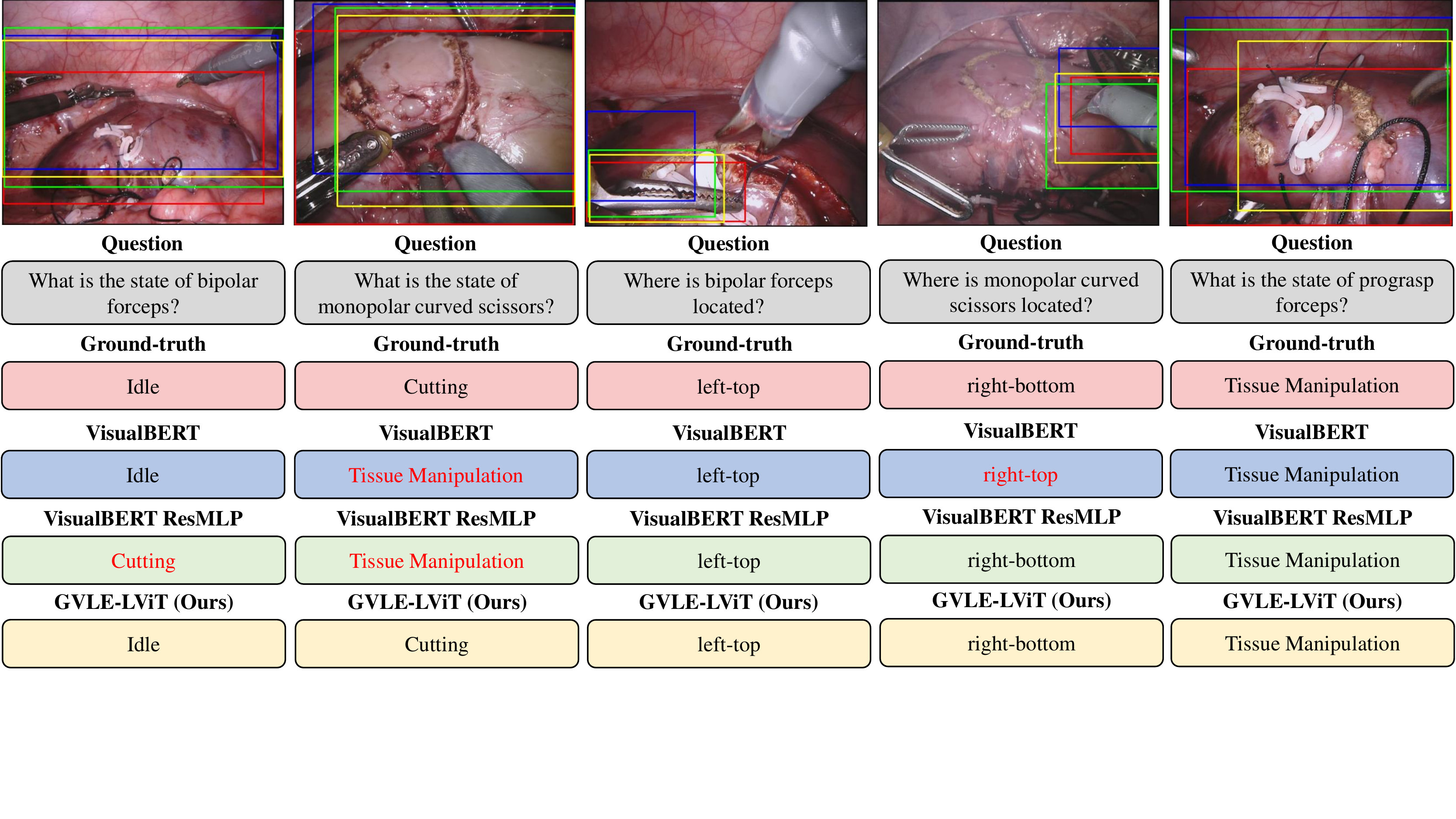}
    \caption{Several examples of answer and bounding box generation by VisualBERT~\cite{li2019visualbert}, VisualBERT ResMLP~\cite{seenivasan2022surgical}, and our GVLE-LViT model. Compared with the baseline models, the localization and classification prediction results of our model are more accurate. The denotation of bounding box color is as follows: red: Ground-truth, blue: VisualBERT~\cite{li2019visualbert}, green: VisualBERT ResMLP~\cite{seenivasan2022surgical}, yellow: GVLE-LViT (Ours).
    }
    \label{fig:qualitative}
\end{figure*}

\begin{table*}[htbp]%
    \renewcommand\arraystretch{0.4}
	\caption{
	    K-fold Comparison experiments of our GVLE-LViT model on VQLA tasks, against VisualBERT~\cite{li2019visualbert} and VisualBERT ResMLP~\cite{seenivasan2022surgical} based model.
	}
	\centering
	\label{tab:2}  
	\setlength{\tabcolsep}{1.5mm}{
	\begin{tabular}{c|c|p{1.0cm}<{\centering} p{1.0cm}<{\centering} p{1.0cm}<{\centering} | p{1.0cm}<{\centering} p{1.0cm}<{\centering} p{1.0cm}<{\centering}}
		\noalign{\smallskip}\hline\noalign{\smallskip}\noalign{\smallskip}	
	\multirow{2}{*}{\makecell[c]{Models}} & \multirow{2}{*}{Fold} & \multicolumn{3}{c}{\textbf{EndoVis-18-VQLA}} & \multicolumn{3}{|c}{\textbf{EndoVis-17-VQLA}  } \\
        \noalign{\smallskip}\cline{3-8}\noalign{\smallskip}
		& & Acc & F-Score & mIoU & Acc & F-Score & mIoU \\\noalign{\smallskip}\hline\noalign{\smallskip}
		VisualBERT~\cite{li2019visualbert} & & 0.6215 & 0.3320 & 0.7356 & 0.3898 & 0.3169 & 0.7105 \\
		\noalign{\smallskip}
		VisualBERT ResMLP~\cite{seenivasan2022surgical} & $1^{st}$ Fold & 0.6320 & 0.3311 & 0.7501 & 0.4195 & \textbf{0.3316} & 0.7035 \\
        \noalign{\smallskip}
        GVLE-LViT (Ours) & & \textbf{0.6659} & \textbf{0.3614} & \textbf{0.7625} & \textbf{0.4576} & 0.2489 & \textbf{0.7275} \\\noalign{\smallskip}\hline\noalign{\smallskip}
		VisualBERT~\cite{li2019visualbert} & & 0.6290 & 0.3458 & 0.7609 & 0.3898 & 0.3333 & 0.7141 \\
		\noalign{\smallskip}
		VisualBERT ResMLP~\cite{seenivasan2022surgical} & $2^{nd}$ Fold & 0.6174 & 0.3365 & 0.7667 & 0.4216 & 0.3787 & 0.7349 \\
        \noalign{\smallskip}
        GVLE-LViT (Ours) & & \textbf{0.6655} & \textbf{0.4122} & \textbf{0.7691} & \textbf{0.4831} & \textbf{0.3953} & \textbf{0.7433} \\\noalign{\smallskip}\hline\noalign{\smallskip}
        VisualBERT~\cite{li2019visualbert} & & 0.5771 & 0.3421 & 0.7440 & \textbf{0.4470} & 0.3488 & 0.7224 \\
		\noalign{\smallskip}
		VisualBERT ResMLP~\cite{seenivasan2022surgical} & $3^{rd}$ Fold & 0.5817 & 0.3794 & 0.7456 & 0.4025 & 0.3271 & 0.7159 \\
        \noalign{\smallskip}
        GVLE-LViT (Ours) & & \textbf{0.6247} & \textbf{0.4062} & \textbf{0.7636} & 0.4449 & \textbf{0.3546} & \textbf{0.7430} \\\noalign{\smallskip}\hline
	\end{tabular}}
\end{table*}

\begin{table*}[t]%
    \renewcommand\arraystretch{0.4}
	\caption{Ablation studies with different localization loss function combinations on our proposed GVLE-LViT model, against VisualBERT~\cite{li2019visualbert} and VisualBERT ResMLP~\cite{seenivasan2022surgical} based model }
	\centering
	\label{tab:3}  
	\setlength{\tabcolsep}{1.5mm}{
	\begin{tabular}{c|p{0.9cm}<{\centering}|p{1.2cm}<{\centering}|p{1.2cm}<{\centering}|p{1.0cm}<{\centering} p{1.0cm}<{\centering} p{1.0cm}<{\centering} | p{1.0cm}<{\centering} p{1.0cm}<{\centering} p{1.0cm}<{\centering}}%
		\noalign{\smallskip}\hline\noalign{\smallskip}\noalign{\smallskip}	
		\multirow{2}{*}{Models} & \multicolumn{3}{c|}{Loss Function} & \multicolumn{3}{c}{\textbf{EndoVis-18-VQLA}} & \multicolumn{3}{|c}{\textbf{EndoVis-17-VQLA}  } \\
        \noalign{\smallskip}\cline{2-10}\noalign{\smallskip}\noalign{\smallskip}	
		& VQA & \multicolumn{2}{c|}{Detection} & Acc & F-Score & mIoU & Acc & F-Score & mIoU \\\noalign{\smallskip}\hline\noalign{\smallskip}
		VisualBERT~\cite{li2019visualbert} & & & & 0.6244 & \textbf{0.3681} & 0.7234 & 0.4174 & \textbf{0.3326} & 0.7136\\\noalign{\smallskip}
		VisualBERT ResMLP~\cite{seenivasan2022surgical} & $\emph{CE}$ & $\mathcal{L}_1$ & $\times$ & 0.6107 & 0.2977 & 0.7383 & 0.3877 & 0.3197 & 0.7089\\
        \noalign{\smallskip}
        GVLE-LViT (Ours) & & & & 0.6287 & 0.2965 & 0.7520 & 0.4407 & 0.2166 & 0.7120\\\noalign{\smallskip}\hline\noalign{\smallskip}
		VisualBERT~\cite{li2019visualbert} & & & & 0.6215 & 0.3320 & 0.7356 & 0.3898 & 0.3169 & 0.7105 \\
		\noalign{\smallskip}
		VisualBERT ResMLP~\cite{seenivasan2022surgical} & $\emph{CE}$ & $\mathcal{L}_1$ & \emph{GIoU}~\cite{rezatofighi2019generalized} & 0.6320 & 0.3311 & 0.7501 & 0.4195 & 0.3316 & 0.7035 \\
        \noalign{\smallskip}
        GVLE-LViT (Ours) & & & & \textbf{0.6659} & 0.3614 & \textbf{0.7625} & \textbf{0.4576} & 0.2489 & \textbf{0.7275} \\
		\noalign{\smallskip}\hline
	\end{tabular}}
\end{table*}

\begin{table*}[h]%
    \renewcommand\arraystretch{0.4}
	\caption{
	    Comparison experiments of our GVLE language-vision embedding fusion against ConCAT~\cite{li2019visualbert}, AFF~\cite{dai2021aff} and iAFF~\cite{dai2021aff} based fusion strategies.} 
	\centering
	\label{tab:4}  
	\setlength{\tabcolsep}{1.5mm}{
	\begin{tabular}{c|p{1cm}<{\centering} p{1cm}<{\centering} p{1cm}<{\centering} | p{1cm}<{\centering} p{1cm}<{\centering} p{1cm}<{\centering}}%
		\noalign{\smallskip}\hline\noalign{\smallskip}\noalign{\smallskip}	
		\multirow{2}{*}{\makecell[c]{Embedding \\ Techniques}} & \multicolumn{3}{c}{\textbf{EndoVis-18-VQLA}} & \multicolumn{3}{|c}{\textbf{EndoVis-17-VQLA}} \\
        \noalign{\smallskip}\cline{2-7}\noalign{\smallskip}\noalign{\smallskip}	
         & Acc & F-Score & mIoU & Acc & F-Score & mIoU \\\noalign{\smallskip}\hline\noalign{\smallskip}
		ConCAT~\cite{li2019visualbert} & 0.6551 & 0.3591 & 0.7386 & 0.4258 & 0.3183 & 0.7035\\
		\noalign{\smallskip}
		AFF~\cite{dai2021aff} & 0.6295 & 0.3521 & 0.7459 & 0.3835 & \textbf{0.3270} & 0.7051\\
        \noalign{\smallskip}
        iAFF~\cite{dai2021aff} & 0.6356 & 0.3339 & 0.7498 & 0.4047 & 0.2948 & 0.7164\\
		\noalign{\smallskip}
        GVLE (Ours) & \textbf{0.6659} & \textbf{0.3614} & \textbf{0.7625} & \textbf{0.4576} & 0.2489 & \textbf{0.7275} \\
		\noalign{\smallskip}\hline
	\end{tabular}}
\end{table*}

The performance of our proposed GVLE-LViT model is both quantitatively (Table~\ref{tab:1}) and qualitatively (Fig.~\ref{fig:qualitative} benchmarked against the state-of-the-art (SOTA) Transformers-based VisualBert~\cite{li2019visualbert} and VisualBert ResMLP~\cite{seenivasan2022surgical} models for Visual Question localized-answering on EndoVis-18-VQLA and EndoVis-17-VQLA dataset. Table~\ref{tab:1} shows that our proposed model outperforms other SOTA models on both datasets. Furthermore, comparing the performance of all three models using the features extracted from the object detection model output against the performance of 3 models using features from the entire image, we note that the performances of the models that use features from the entire image are consistently superior. This superior performance can be attributed to the model’s ability to perform global scene understanding from the complete image (in-line with the observation made by Seenivasan et.al~\cite{seenivasan2022global}) and optimal convergence of the model from end-to-end model training. Additionally, by removing the need for an object detection network, we increased the model’s processing speed by more than $8$ times, achieving $150.6$ frames per second (FPS) and making it suitable for real-time applications. Qualitatively, as shown in Fig.~\ref{fig:qualitative}, our model outperforms base models in both answering and localization (close to ground-truth bounding box annotation).

A K-fold study is also conducted to study and prove our model’s superiority over the base model. We set up three different ways of splitting the training and test set. Table~\ref{tab:2} proves that our proposed GVLE-LViT model generally outperforms the base Transformer-based models on all three folds on both datasets. 

\subsection{Ablation Studies}
\label{sec:3.4}
Firstly, an ablation study on the performance of the models trained using different loss function combinations is studied (Table~\ref{tab:3}). As our GVLE-LViT and transformed-based baseline models (VisualBERT~\cite{li2019visualbert} and VisualBERT ResMLP~\cite{seenivasan2022surgical}) aim to predict the localized answer, the loss function for both answer prediction and answer location is used during the training process. From Table~\ref{tab:3}, it is observed that in addition to cross-entropy (CE) loss (for answer prediction) and $\mathcal{L}_1$ loss (for answer localization), incorporating GIoU~\cite{rezatofighi2019generalized} loss (for answer localization) significantly improves the model’s performance in both answer prediction and answer localization.

Secondly, an ablation study on various techniques of heterogenous feature fusion is studied. Our proposed GVLE feature fusion technique is compared against ConCAT~\cite{li2019visualbert}, AFF~\cite{dai2021aff} and iAFF~\cite{dai2021aff} techniques. Table~\ref{tab:4} proves that our proposed GVLE vision-language feature fusion technique outperforms other feature fusion techniques.

\section{CONCLUSIONS}
\label{sec:4}

We design and propose a Surgical Visual Question Localized-Answering (Surgical-VQLA) model that can answer “what?” and “where?” based on a given input question and surgical scene, making it easier for the student to infer “why?”. Specifically, we propose a GVLE-LViT model that better fuses heterogeneous features (visual and text) using our proposed GVLE technique that outperforms existing SOTA models in Surgical-VQLA tasks on two surgical datasets. Additionally, we integrate GIoU loss with cross-entropy loss and $\mathcal{L}_1$ loss to improve both the prediction and localization performance of the model. 
Through extensive comparative, k-fold and ablation studies, we prove that our proposed GVLE-LViT trained using our proposed loss combination outperforms existing SOTA models. The Surgical-VQLA system may become an important auxiliary tool in surgical training.

While the proposed VQLA model aims to provide reliable answer prediction, to an extent, the localization of the answer could help quantify the reliability of prediction on new data, where if the localization is far-off than the target instrument or tissue, the user can infer that the prediction is probably wrong or the input data is out-of-distribution data. 
Therefore, using localization information to predict prediction reliability could be a possible future work. In an application stance, our proposed VQLA model opens novel possible applications for medical diagnosis. More complicated datasets and challenging QA pairs shall further boost the prospective of the Surgical-VQLA system.







\balance
\bibliographystyle{IEEEtran}
\bibliography{reference}

\begin{thebibliography}{10}
\providecommand{\url}[1]{#1}
\csname url@samestyle\endcsname
\providecommand{\newblock}{\relax}
\providecommand{\bibinfo}[2]{#2}
\providecommand{\BIBentrySTDinterwordspacing}{\spaceskip=0pt\relax}
\providecommand{\BIBentryALTinterwordstretchfactor}{4}
\providecommand{\BIBentryALTinterwordspacing}{\spaceskip=\fontdimen2\font plus
\BIBentryALTinterwordstretchfactor\fontdimen3\font minus
  \fontdimen4\font\relax}
\providecommand{\BIBforeignlanguage}[2]{{%
\expandafter\ifx\csname l@#1\endcsname\relax
\typeout{** WARNING: IEEEtran.bst: No hyphenation pattern has been}%
\typeout{** loaded for the language `#1'. Using the pattern for}%
\typeout{** the default language instead.}%
\else
\language=\csname l@#1\endcsname
\fi
#2}}
\providecommand{\BIBdecl}{\relax}
\BIBdecl

\bibitem{sharma2021medfusenet}
D.~Sharma, S.~Purushotham, and C.~K. Reddy, ``Medfusenet: An attention-based
  multimodal deep learning model for visual question answering in the medical
  domain,'' \emph{Scientific Reports}, vol.~11, no.~1, pp. 1--18, 2021.

\bibitem{seenivasan2022surgical}
L.~Seenivasan, M.~Islam, A.~Krishna, and H.~Ren, ``Surgical-vqa: Visual
  question answering in surgical scenes using transformer,'' in
  \emph{International Conference on Medical Image Computing and
  Computer-Assisted Intervention}.\hskip 1em plus 0.5em minus 0.4em\relax
  Springer, 2022, pp. 33--43.

\bibitem{hsieh2017vr}
M.-C. Hsieh and Y.-H. Lin, ``Vr and ar applications in medical practice and
  education,'' \emph{Hu Li Za Zhi}, vol.~64, no.~6, pp. 12--18, 2017.

\bibitem{ashraf2018eye}
H.~Ashraf, M.~H. Sodergren, N.~Merali, G.~Mylonas, H.~Singh, and A.~Darzi,
  ``Eye-tracking technology in medical education: A systematic review,''
  \emph{Medical teacher}, vol.~40, no.~1, pp. 62--69, 2018.

\bibitem{lin2006towards}
H.~C. Lin, I.~Shafran, D.~Yuh, and G.~D. Hager, ``Towards automatic skill
  evaluation: Detection and segmentation of robot-assisted surgical motions,''
  \emph{Computer Aided Surgery}, vol.~11, no.~5, pp. 220--230, 2006.

\bibitem{che2022learning}
H.~Che, H.~Jin, and H.~Chen, ``Learning robust representation for joint grading
  of ophthalmic diseases via adaptive curriculum and feature disentanglement,''
  in \emph{Medical Image Computing and Computer Assisted Intervention--MICCAI
  2022: 25th International Conference, Singapore, September 18--22, 2022,
  Proceedings, Part III}.\hskip 1em plus 0.5em minus 0.4em\relax Springer,
  2022, pp. 523--533.

\bibitem{bai2021influence}
L.~Bai, S.~Chen, M.~Gao, L.~Abdelrahman, M.~Al~Ghamdi, and M.~Abdel-Mottaleb,
  ``The influence of age and gender information on the diagnosis of diabetic
  retinopathy: based on neural networks,'' in \emph{2021 43rd annual
  international conference of the IEEE engineering in medicine \& Biology
  Society (EMBC)}.\hskip 1em plus 0.5em minus 0.4em\relax IEEE, 2021, pp.
  3514--3517.

\bibitem{bai2022transformer}
L.~Bai, L.~Wang, T.~Chen, Y.~Zhao, and H.~Ren, ``Transformer-based disease
  identification for small-scale imbalanced capsule endoscopy dataset,''
  \emph{Electronics}, vol.~11, no.~17, p. 2747, 2022.

\bibitem{li2019visualbert}
L.~H. Li, M.~Yatskar, D.~Yin, C.-J. Hsieh, and K.-W. Chang, ``Visualbert: A
  simple and performant baseline for vision and language,'' \emph{arXiv
  preprint arXiv:1908.03557}, 2019.

\bibitem{sheng2021human}
S.~Sheng, A.~Singh, V.~Goswami, J.~Magana, T.~Thrush, W.~Galuba, D.~Parikh, and
  D.~Kiela, ``Human-adversarial visual question answering,'' \emph{Advances in
  Neural Information Processing Systems}, vol.~34, 2021.

\bibitem{wang2021simvlm}
Z.~Wang, J.~Yu, A.~W. Yu, Z.~Dai, Y.~Tsvetkov, and Y.~Cao, ``Simvlm: Simple
  visual language model pretraining with weak supervision,'' \emph{arXiv
  preprint arXiv:2108.10904}, 2021.

\bibitem{sharma2021image}
H.~Sharma and A.~S. Jalal, ``Image captioning improved visual question
  answering,'' \emph{Multimedia Tools and Applications}, pp. 1--22, 2021.

\bibitem{barra2021visual}
S.~Barra, C.~Bisogni, M.~De~Marsico, and S.~Ricciardi, ``Visual question
  answering: which investigated applications?'' \emph{Pattern Recognition
  Letters}, vol. 151, pp. 325--331, 2021.

\bibitem{sharma2021visual}
H.~Sharma and A.~S. Jalal, ``Visual question answering model based on graph
  neural network and contextual attention,'' \emph{Image and Vision Computing},
  vol. 110, p. 104165, 2021.

\bibitem{seenivasan2022global}
L.~Seenivasan, S.~Mitheran, M.~Islam, and H.~Ren, ``Global-reasoned multi-task
  learning model for surgical scene understanding,'' \emph{IEEE Robotics and
  Automation Letters}, 2022.

\bibitem{dai2021aff}
Y.~Dai, F.~Gieseke, S.~Oehmcke, Y.~Wu, and K.~Barnard, ``Attentional feature
  fusion,'' in \emph{Proceedings of the IEEE/CVF Winter Conference on
  Applications of Computer Vision}, 2021, pp. 3560--3569.

\bibitem{rezatofighi2019generalized}
H.~Rezatofighi, N.~Tsoi, J.~Gwak, A.~Sadeghian, I.~Reid, and S.~Savarese,
  ``Generalized intersection over union: A metric and a loss for bounding box
  regression,'' in \emph{Proceedings of the IEEE/CVF conference on computer
  vision and pattern recognition}, 2019, pp. 658--666.

\bibitem{devlin2018bert}
J.~Devlin, M.-W. Chang, K.~Lee, and K.~Toutanova, ``Bert: Pre-training of deep
  bidirectional transformers for language understanding,'' \emph{arXiv preprint
  arXiv:1810.04805}, 2018.

\bibitem{wu2016google}
Y.~Wu, M.~Schuster, Z.~Chen, Q.~V. Le, M.~Norouzi, W.~Macherey, M.~Krikun,
  Y.~Cao, Q.~Gao, K.~Macherey \emph{et~al.}, ``Google's neural machine
  translation system: Bridging the gap between human and machine translation,''
  \emph{arXiv preprint arXiv:1609.08144}, 2016.

\bibitem{touvron2021resmlp}
H.~Touvron, P.~Bojanowski, M.~Caron, M.~Cord, A.~El-Nouby, E.~Grave,
  G.~Izacard, A.~Joulin, G.~Synnaeve, J.~Verbeek \emph{et~al.}, ``Resmlp:
  Feedforward networks for image classification with data-efficient training,''
  \emph{arXiv preprint arXiv:2105.03404}, 2021.

\bibitem{dosovitskiy2020image}
A.~Dosovitskiy, L.~Beyer, A.~Kolesnikov, D.~Weissenborn, X.~Zhai,
  T.~Unterthiner, M.~Dehghani, M.~Minderer, G.~Heigold, S.~Gelly \emph{et~al.},
  ``An image is worth 16x16 words: Transformers for image recognition at
  scale,'' \emph{arXiv preprint arXiv:2010.11929}, 2020.

\bibitem{vaswani2017attention}
A.~Vaswani, N.~Shazeer, N.~Parmar, J.~Uszkoreit, L.~Jones, A.~N. Gomez,
  {\L}.~Kaiser, and I.~Polosukhin, ``Attention is all you need,''
  \emph{Advances in neural information processing systems}, vol.~30, 2017.

\bibitem{he2016resnet}
K.~He, X.~Zhang, S.~Ren, and J.~Sun, ``Deep residual learning for image
  recognition,'' in \emph{Proceedings of the IEEE conference on computer vision
  and pattern recognition}, 2016, pp. 770--778.

\bibitem{deng2009imagenet}
J.~Deng, W.~Dong, R.~Socher, L.-J. Li, K.~Li, and L.~Fei-Fei, ``Imagenet: A
  large-scale hierarchical image database,'' in \emph{2009 IEEE conference on
  computer vision and pattern recognition}.\hskip 1em plus 0.5em minus
  0.4em\relax Ieee, 2009, pp. 248--255.

\bibitem{srivastava2012multimodal}
N.~Srivastava and R.~R. Salakhutdinov, ``Multimodal learning with deep
  boltzmann machines,'' \emph{Advances in neural information processing
  systems}, vol.~25, 2012.

\bibitem{arevalo2017gmu}
J.~Arevalo, T.~Solorio, M.~Montes-y G{\'o}mez, and F.~A. Gonz{\'a}lez, ``Gated
  multimodal units for information fusion,'' \emph{arXiv preprint
  arXiv:1702.01992}, 2017.

\bibitem{allan2020endovis18}
M.~Allan, S.~Kondo, S.~Bodenstedt, S.~Leger, R.~Kadkhodamohammadi, I.~Luengo,
  F.~Fuentes, E.~Flouty, A.~Mohammed, M.~Pedersen \emph{et~al.}, ``2018 robotic
  scene segmentation challenge,'' \emph{arXiv preprint arXiv:2001.11190}, 2020.

\bibitem{islam2020learning}
M.~Islam, L.~Seenivasan, L.~C. Ming, and H.~Ren, ``Learning and reasoning with
  the graph structure representation in robotic surgery,'' in
  \emph{International Conference on Medical Image Computing and
  Computer-Assisted Intervention}.\hskip 1em plus 0.5em minus 0.4em\relax
  Springer, 2020, pp. 627--636.

\bibitem{allan2019endovis17}
M.~Allan, A.~Shvets, T.~Kurmann, Z.~Zhang, R.~Duggal, Y.-H. Su, N.~Rieke,
  I.~Laina, N.~Kalavakonda, S.~Bodenstedt \emph{et~al.}, ``2017 robotic
  instrument segmentation challenge,'' \emph{arXiv preprint arXiv:1902.06426},
  2019.

\bibitem{kingma2014adam}
D.~P. Kingma and J.~Ba, ``Adam: A method for stochastic optimization,''
  \emph{arXiv preprint arXiv:1412.6980}, 2014.

\bibitem{ren2015faster}
S.~Ren, K.~He, R.~Girshick, and J.~Sun, ``Faster r-cnn: Towards real-time
  object detection with region proposal networks,'' \emph{Advances in neural
  information processing systems}, vol.~28, 2015.

\end{thebibliography}

\end{document}